\def\year{2018}\relax
\documentclass[letterpaper]{article} 
\usepackage{aaai18}  
\usepackage{times}  
\usepackage{helvet}  
\usepackage{courier}  
\usepackage{url}  
\usepackage{graphicx}  
\usepackage{xcolor}
\usepackage{nameref}
\usepackage{subfigure}
\usepackage{algorithm, algorithmic}
\usepackage{amsmath, amsthm, amssymb}
\usepackage{arydshln}
\usepackage{listings,xcolor}

\theoremstyle{remark}

\begin{document}
%

\title{MAgent: A Many-Agent Reinforcement Learning Platform \\for Artificial Collective Intelligence}
\author{
   Lianmin Zheng{$^{\dag}$}, Jiacheng Yang{$^{\dag}$\thanks{The first two authors have equal contributions. Correspondence to Weinan Zhang, wnzhang@sjtu.edu.cn, and Jun Wang, j.wang@cs.ucl.ac.uk.}~}, Han Cai{$^\dag$},
   Weinan Zhang{$^\dag$}, Jun Wang{$^\ddag$}, Yong Yu{$^{\dag}$} \\
    {$^\dag$}Shanghai Jiao Tong University, {$^\ddag$}University College London\\
}
\maketitle

\begin{abstract}
We introduce MAgent, a platform to support research and development of many-agent reinforcement learning. 
Unlike previous research platforms on single or multi-agent reinforcement learning, MAgent focuses on supporting the tasks and the applications that require hundreds to millions of agents.
Within the interactions among a population of agents, it enables not only the study of learning algorithms for agents' optimal polices, but more importantly, the observation and understanding of individual agent's behaviors and social phenomena emerging from the AI society, including communication languages, leaderships, altruism.
MAgent is highly scalable and can host up to one million agents on a single GPU server. MAgent also provides flexible configurations for AI researchers to design their customized environments and agents. In this demo, we present three environments designed on MAgent and show emerged collective intelligence by learning from scratch.
\end{abstract}

\section{Introduction}

True human intelligence embraces social and collective wisdom, 
laying a foundation for general Artificial Intelligence (AI)~\cite{goertzel2007artificial}. In essence, human intelligence would collectively solve the problem that otherwise is unthinkable by a single person. For instance, theoretically, Condorcet's jury theorem \cite{landemore2013democratic} shows that, under certain assumption, adding more voters in a group would increase the probability that the majority chooses the right answer. 

In parallel, in the coming era of algorithmic economy, many AI agents work together, \emph{artificially} creating their own collective intelligence. With certain rudimentary abilities, {Artificial Collective Intelligence} (ACI) starts to emerge from multiple domains, including stock trading \cite{wang2013investigating}, strategic games \cite{peng2017multiagent} and city transportation optimization \cite{zhang2017taxi} etc.

A key technology of ACI is multi-agent reinforcement learning (RL) but typically requires a scale of hundreds to millions of agents \cite{1magent}. However, existing experimentation platforms, including ALE \cite{bellemare2013the}, OpenAI Gym/Universe \cite{brockman2016openai}, Malmo \cite{johnson2016malmo}, ELF \cite{tian2017elf} and SC2LE \cite{vinyals2017starcraft} fail to meet the demand. Although they gradually show the desire to cover multi-agent scenarios, they are basically designed to take no more than dozens of agents. Thus there is a significant need for a platform dedicated to large population multi-agent reinforcement learning, which is critical for ACI. On the other hand, the majority of state-of-the-art multi-agent reinforcement learning algorithms \cite{lowe2017multi} are also limited in the scale of dozens of agents. Therefore, it is also a great challenge to the research community.



\section{The MAgent Platform}

The MAgent project\footnote{https://github.com/geek-ai/MAgent}  aims to build a many-agent reinforcement learning platform for the research of ACI. With the idea of network sharing and ID embedding, MAgent is highly scalable and can host up to one million agents on a single GPU server. 
Moreover, MAgent provides environment/agent configurations and a reward description language to enable flexible environment design. Finally, MAgent provides a simple yet visually effective render to interactively present the state of environment and agents. Users can also slide or zoom the scope window and even manipulate agents in the game.



\begin{figure*}[t]
	\centering
	\vspace{-10pt}
	\subfigure[Pursuit (1)]{
		\includegraphics[width=0.18\textwidth]{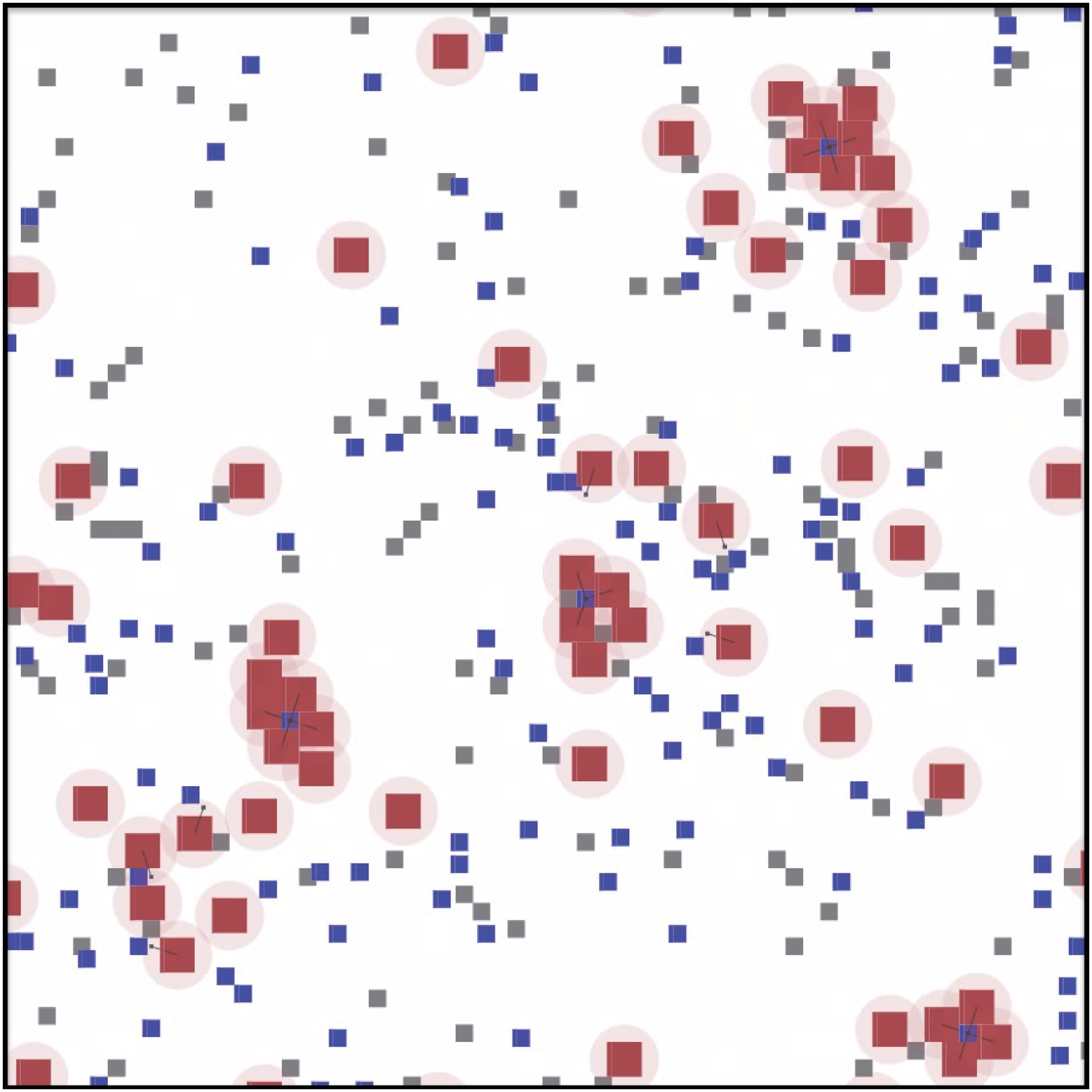}\label{fig:pursuit1}
	}
	\subfigure[Pursuit (2)]{
		\includegraphics[width=0.18\textwidth]{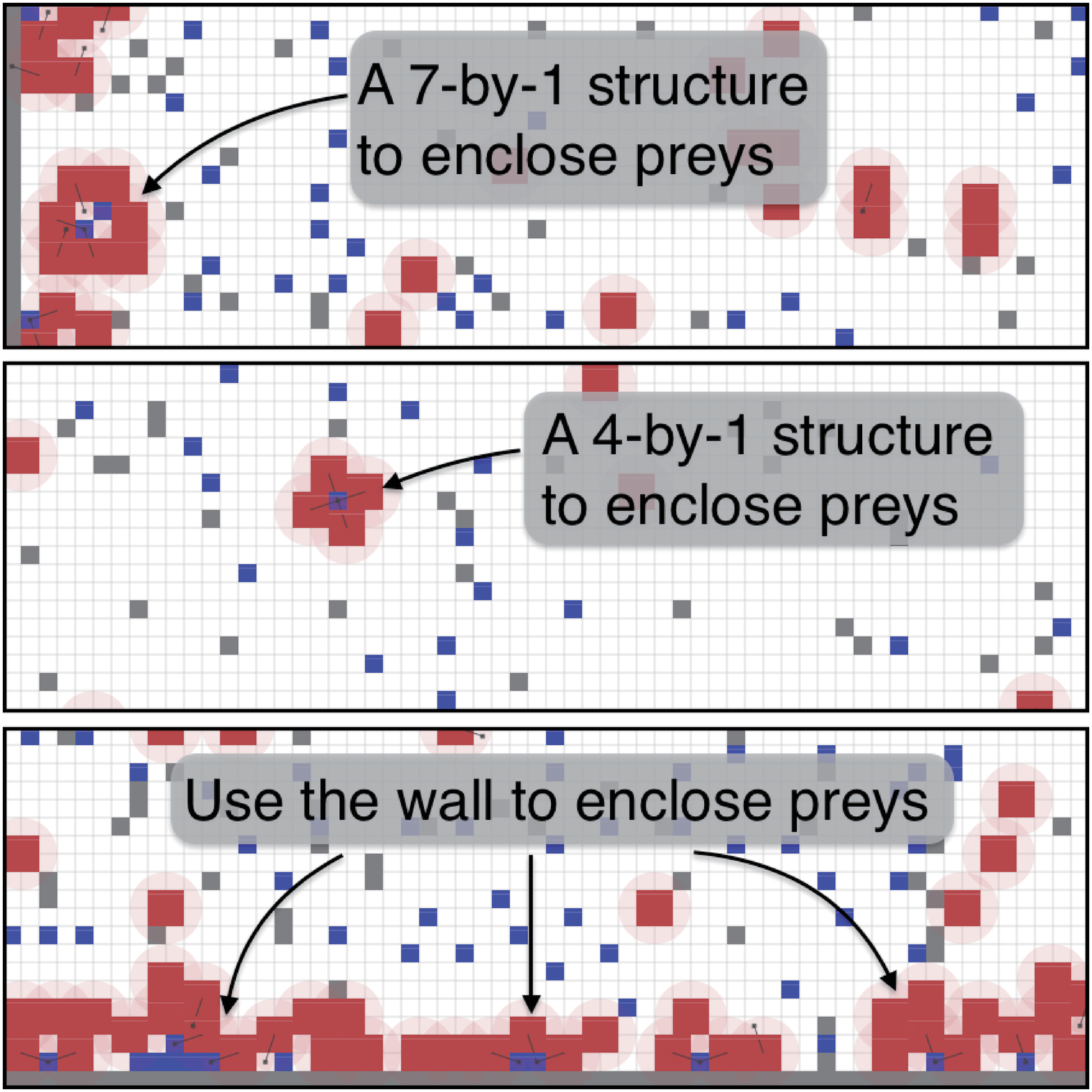}\label{fig:pursuit2}
	}
	\subfigure[Gathering]{
		\includegraphics[width=0.18\textwidth]{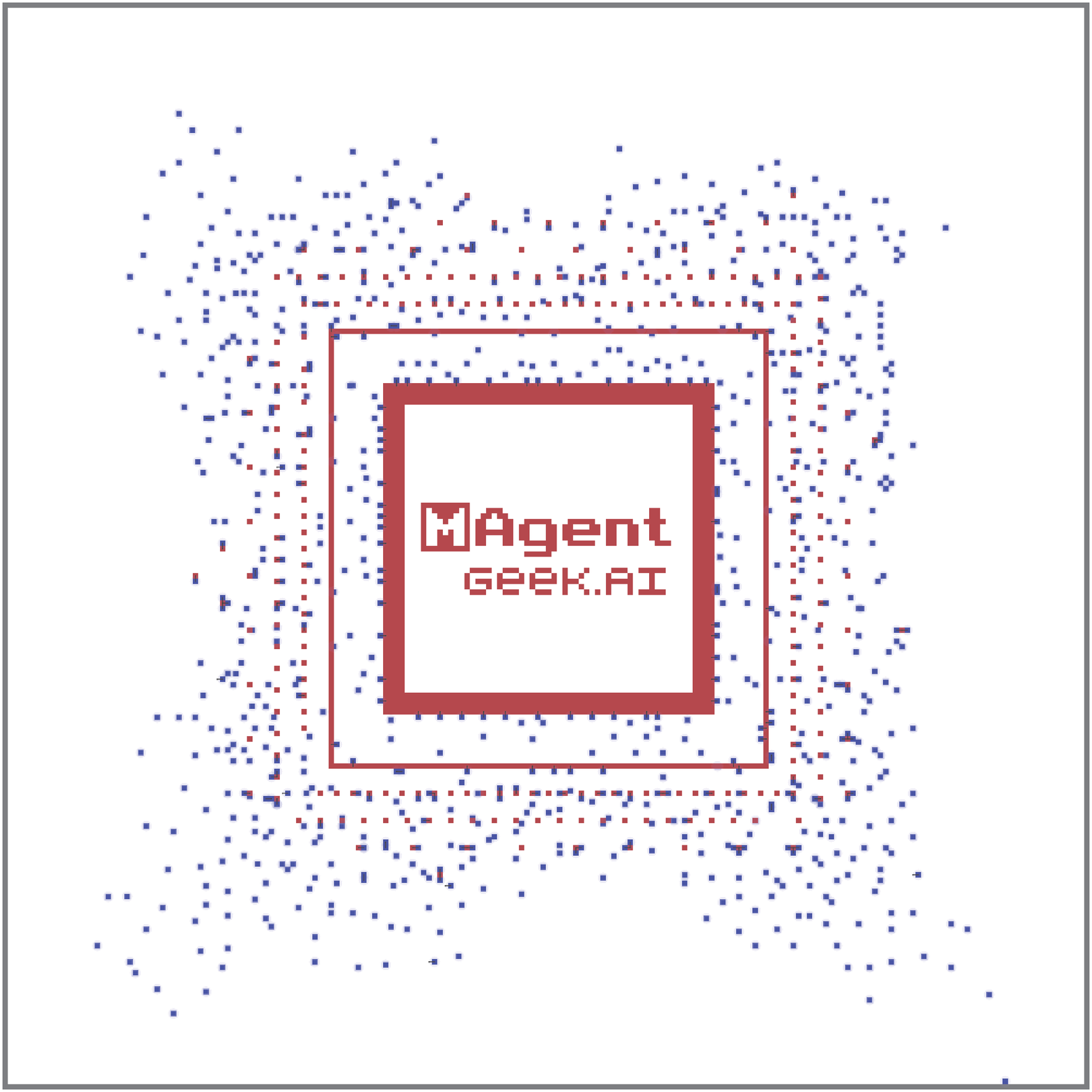}\label{fig:gather}
	}
	\subfigure[Battle (1)]{
		\includegraphics[width=0.18\textwidth]{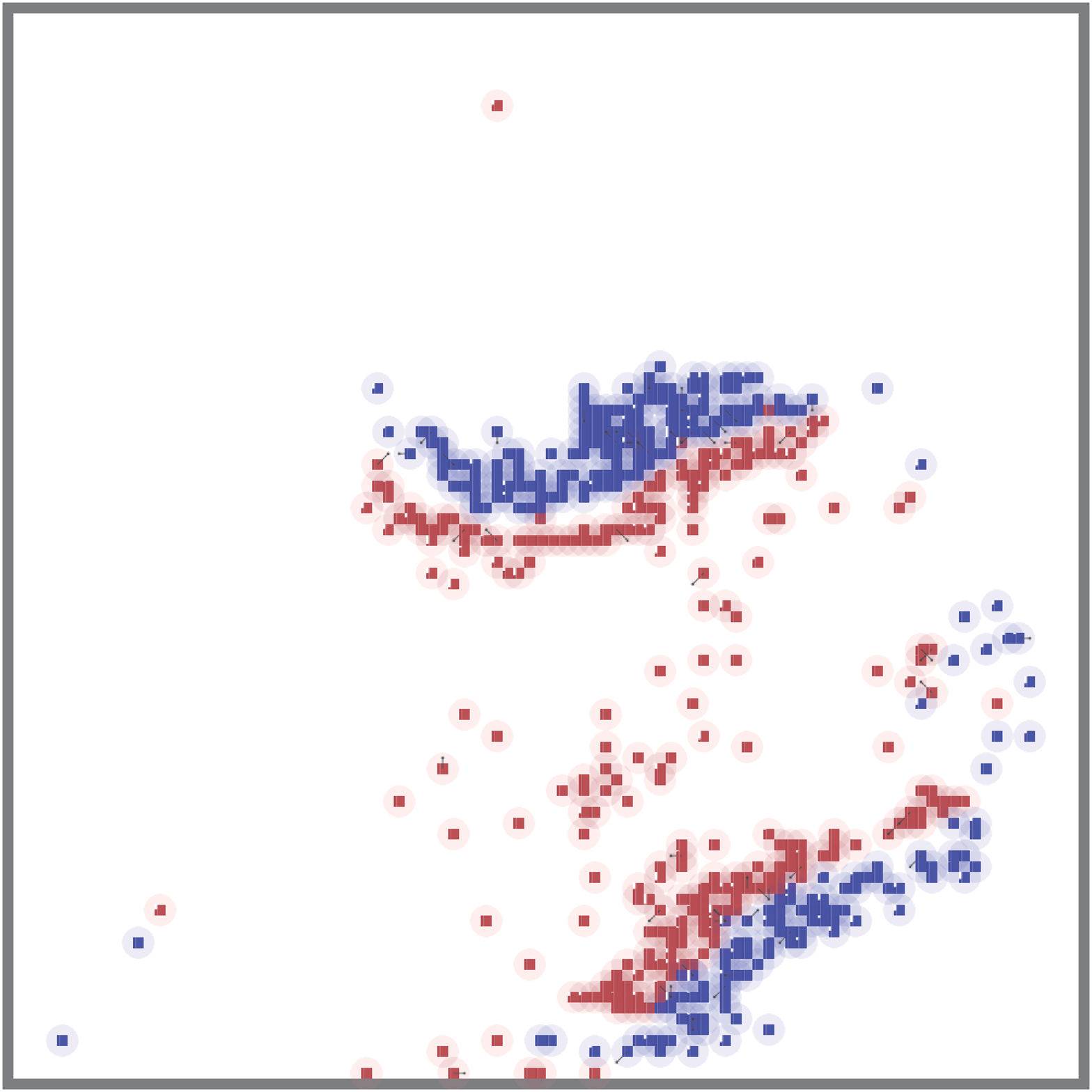}\label{fig:battle1}
	}
	\subfigure[Battle (2)]{
		\includegraphics[width=0.18\textwidth]{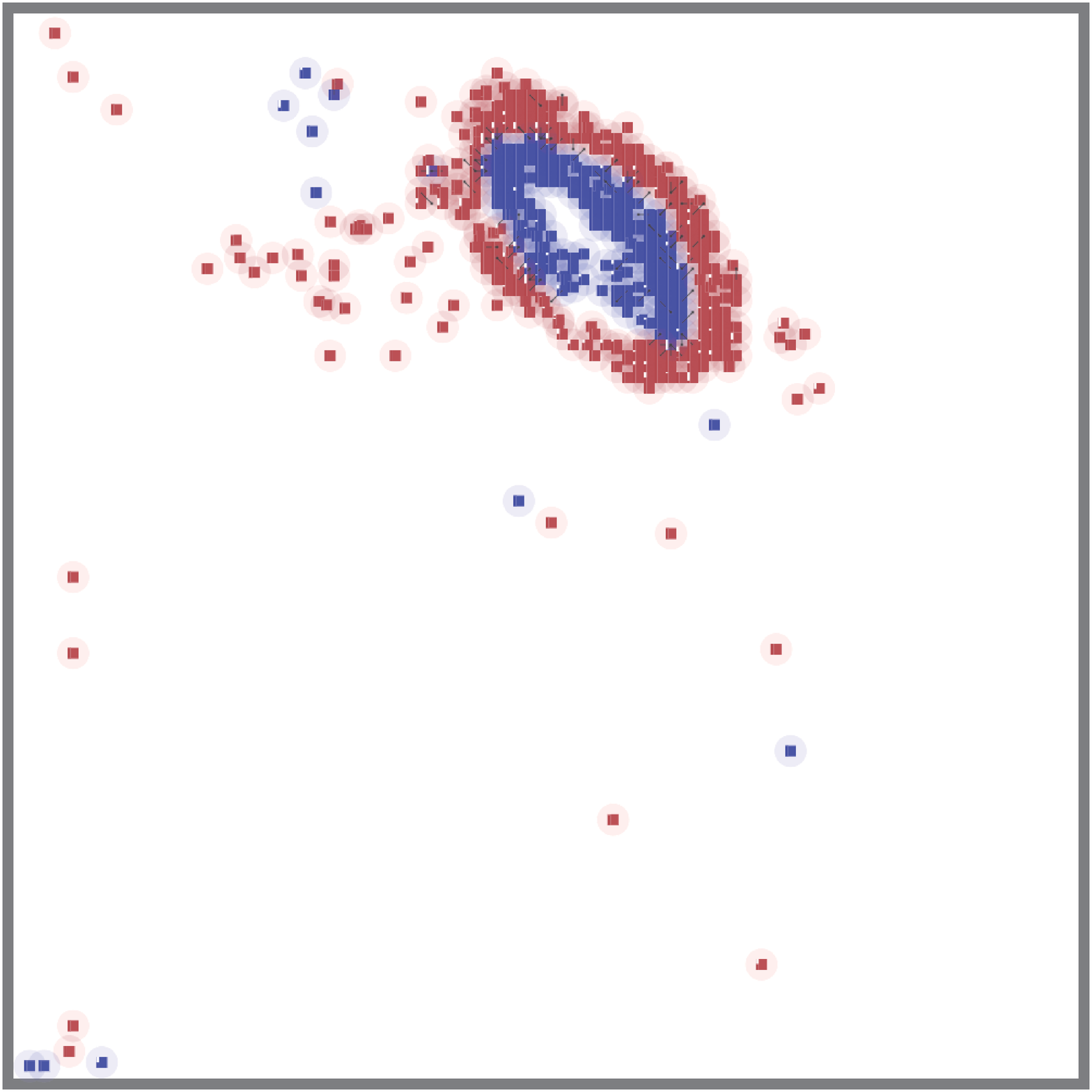}\label{fig:bettle2}
	}
	\vspace{-10pt}
	\caption{Illustrations of the three running examples. }\label{figure:games}\vspace{-10pt}
\end{figure*}

\subsection{Gridworld Fundamentals}
A large-scale gridworld is used as the fundamental environment for the large population of agents. Each agent has a rectangular body with a local detailed perspective and (optional) global information. The actions could be moving, turning or attacking. A C++ engine is provided to support fast simulation. Heterogeneous agents can be registered in our engine. The state space, action space and agents' attributes can be configured easily so that various environments can be swiftly developed.  

\subsection{Reward Description Language}

In our python interface, users can describe the reward, agent symbols, and events by a description language we developed. When the boolean expression of an event expression is true, reward will be assigned to the agents involved in the events.
Logical operation like `and', `or', and `not' is also supported.
The following is an example used to describe the rule of game \emph{pursuit} as described in Live and Interactive Part Section.

\lstset{basicstyle=\scriptsize\sffamily}

\begin{lstlisting}
    a = AgentSymbol(g_predator, index=`any')
    b = AgentSymbol(g_prey, index=`any')
    add_reward_rule(on=Event(a, `attack', b),
                    receiver=[a, b], value=[1, -1])
\end{lstlisting}

\subsection{Baseline Algorithms}
We implement parameter-sharing DQN, DRQN and A2C in our platform. We found DQN performs best in our settings and mainly use it to do following experiment. To bring in the diversity of agents, we also train an embedding for each agent's unique ID. Users can benchmark their own  multi-agent algorithms against these algorithms.


\section{Live and Interactive Part}
In the demo, we will show some example games conducted on MAgent. Demo visitors can see what will happen when applying deep reinforcement learning for such a large number of agents. 
They can use our render to explore in the gridworld, find intelligent patterns and the diversity of RL agents. Three examples are available at present, namely \emph{pursuit}, \emph{gathering} and \emph{battle}. Some illustrations are shown in Fig.~\ref{figure:games}.

\begin{description}

\item [Pursuit\label{item:pursuit}]
shows the emergence of local cooperation. Predators can get rewards if they attack preys while preys will get penalties if they are attacked. After training, predators learn to cooperate with nearby teammates to form several types of closure to lock the preys (see Fig.~\ref{fig:pursuit2}), by which they can get reward every step afterward.

\item [Gathering]
shows the emergence of competition in a limited resource environment. Agents are in a dilemma whether to get reward directly by eating food or killing other agents to monopolize the food. After training, they learn to rush to eat food at first. But when two agents come close, they will try to kill each other.

\item [Battle]
shows the hybrid of cooperation and competition. There are two armies in the map, each of which consists of hundreds of agents. The goal is to collaborate with teammates and eliminate all opponent agents. Simple self-play training enables them to learn both global and local strategies, including encirclement attack, guerrilla warfare.
\end{description}

In addition, we maintain a human player interface. Thus our demo visitors will be able to place themselves in the large world by controlling several agents to gain reward by cooperating or competing with RL agents.

\section{Conclusion}
MAgent is a research platform designed to support many-agent RL. With MAgent, researchers can study a population of AI agents in both individual agent and society levels. For future work, we would support continuous environments and provide more algorithms in MAgent.

{\small
\bibliographystyle{aaai}
\bibliography{magent}
}

\end{document}